\documentclass[runningheads]{llncs}
\usepackage{amsmath}
\usepackage{amsfonts}
\usepackage{bbm}
\usepackage{array}
\newcolumntype{L}{>{\raggedright\arraybackslash}X}
\newcolumntype{C}{>{\centering\arraybackslash}X}
\usepackage{tabularx}
\usepackage{booktabs}
\usepackage{multirow}
\usepackage[table]{xcolor}
\usepackage{xcolor}
\usepackage{threeparttable}

\usepackage[T1]{fontenc}
\usepackage{graphicx,verbatim}
\usepackage{threeparttable}
\begin{document}
\title{PreSight: Preoperative Outcome Prediction for Parkinson’s Disease via
Region-Prior Morphometry and Patient-Specific Weighting}
\titlerunning{PreSight: Preoperative PD Outcome Prediction}
%

\author{
YAN WANG\inst{1} \and
Chen Zhang\inst{1} \and
Lanyun Zhu\inst{2} \and
Yixin Chen\inst{3} \and
Qunbo Wang\inst{1} \and
Yutong Bai\inst{4} \and
Jurgen Germann\inst{6} \and
Yinghong Wen\inst{1} \and
Shuai Shao\inst{7}
}

\authorrunning{Wang et al.}
\institute{
Beijing Jiaotong University\inst{1} \and
Nanyang Technological University\inst{2} \and
Institute of Medical Technology, Peking University\inst{3} \and
Beijing Tiantan Hospital, Capital Medical University\inst{4} \and
University Health Network, University of Toronto\inst{6} \and
Suzhou Institute for Advanced Research, University of Science and Technology of China\inst{7}\\
\email{
yanwang2@bjtu.edu.cn
}
}
\maketitle              
\begin{abstract}
Preoperative improvement rate prediction for Parkinson’s disease surgery is clinically important yet difficult because imaging signals are subtle and patients are heterogeneous. We address this setting, where only information available before surgery is used, and the goal is to predict patient-specific postoperative motor benefit. We present PreSight, a presurgical outcome model that fuses clinical priors with preoperative MRI and deformation-based morphometry (DBM) and adapts regional importance through a patient-specific weighting module. The model produces end-to-end, calibrated, decision-ready predictions with patient-level explanations. We evaluate PreSight on a real-world two-center cohort of 400 subjects with multimodal presurgical inputs and postoperative improvement labels. PreSight outperforms strong clinical, imaging-only, and multimodal baselines. It attains 88.89\% accuracy on internal validation and 85.29\% on an external-center test for responder classification and shows better probability calibration and higher decision-curve net benefit. Ablations and analyses confirm the contribution of DBM and the patient-specific weighting module and indicate that the model emphasizes disease-relevant regions in a patient-specific manner. These results demonstrate that integrating clinical prior knowledge with region-adaptive morphometry enables reliable presurgical decision support in routine practice.

\end{abstract}

\section{Introduction}

Parkinson’s disease (PD) is a progressive neurodegenerative disorder affecting more than 8.5 million individuals worldwide \cite{WHO2022PDBrief}. Deep brain stimulation (DBS) targeting the subthalamic nucleus is an established surgical therapy for advanced PD \cite{limousin2019long,Weaver2009JAMA,Schuepbach2013NEJM}, yet postoperative motor benefit varies substantially across patients. Given the risks, cost, and lifelong management of implanted devices, accurately identifying likely responders before surgery is essential for candidacy selection and shared decision-making.

DBS management spans preoperative evaluation, surgical implantation, and postoperative programming with long-term follow-up. Clinical benefit is commonly quantified by improvement in MDS--UPDRS~III motor scores \cite{Goetz2008MDSUPDRS}. The clinically relevant task is therefore to predict, \emph{using only preoperative information}, whether a patient will achieve meaningful postoperative improvement. Unlike diagnostic imaging tasks, this setting requires forecasting therapeutic response under incomplete and heterogeneous data, making it both clinically critical and methodologically challenging.


Recent advances in clinical prediction modeling emphasize rigorous validation and calibration, as reflected in TRIPOD+AI and PROBAST+AI guidelines \cite{Collins2024BMJ,Moons2025BMJ}. In PD/DBS research, connectomic analyses demonstrate that effective stimulation sites share characteristic network fingerprints \cite{Horn2017AnnNeurol,Li2020NatCommun}, and multimodal learning approaches increasingly integrate clinical scales and imaging features to predict outcomes \cite{Chang2024Frontiers,Biesheuvel2025DBSJ}. However, most prior studies either incorporate postoperative information, focus on population-level associations, or treat brain regions uniformly without adapting to individual clinical context. Preoperative DBS outcome prediction remains difficult for two reasons. First, structural MRI in PD exhibits subtle, spatially distributed morphometric changes, lacking high-contrast biomarkers. Whole-brain modeling can dilute informative signals from disease-relevant circuits. Second, PD patients are highly heterogeneous in stage, symptom profile, and progression rate.

To address this, we propose \textbf{PreSight}, a patient-specific preoperative outcome prediction framework that fuses structural MRI with deformation-based morphometry (DBM) and clinical information. Consistent with prior DBS clinical trials and outcome prediction studies\cite{wolke2023role}, we define responders as patients achieving $\geq$30\% improvement in MDS--UPDRS~III motor scores and formulate the problem as binary responder classification. DBM features derived from T1-weighted MRI capture subtle regional morphometric variation. A central component is a \textbf{Patient-Specific Weighting Module (PSWM)}, which conditions regional importance on individual clinical profiles. Inspired by conditioning and gating mechanisms in multimodal learning \cite{Perez2018AAAI,Shazeer2017ICLR}, PSWM adaptively modulates imaging features rather than applying fixed global weights, enabling individualized decision-ready predictions. We evaluate PreSight on a two-center cohort of 400 PD patients with multimodal preoperative data and postoperative outcomes. Across internal validation and external testing, PreSight improves responder classification while maintaining favorable calibration and decision-curve net benefit. In addition to predictive performance, the model produces interpretable regional weighting patterns aligned with PD-relevant neuroanatomy.

\section{Methodology}
\begin{figure*}[t]
    \centering
    \includegraphics[width=1\linewidth]{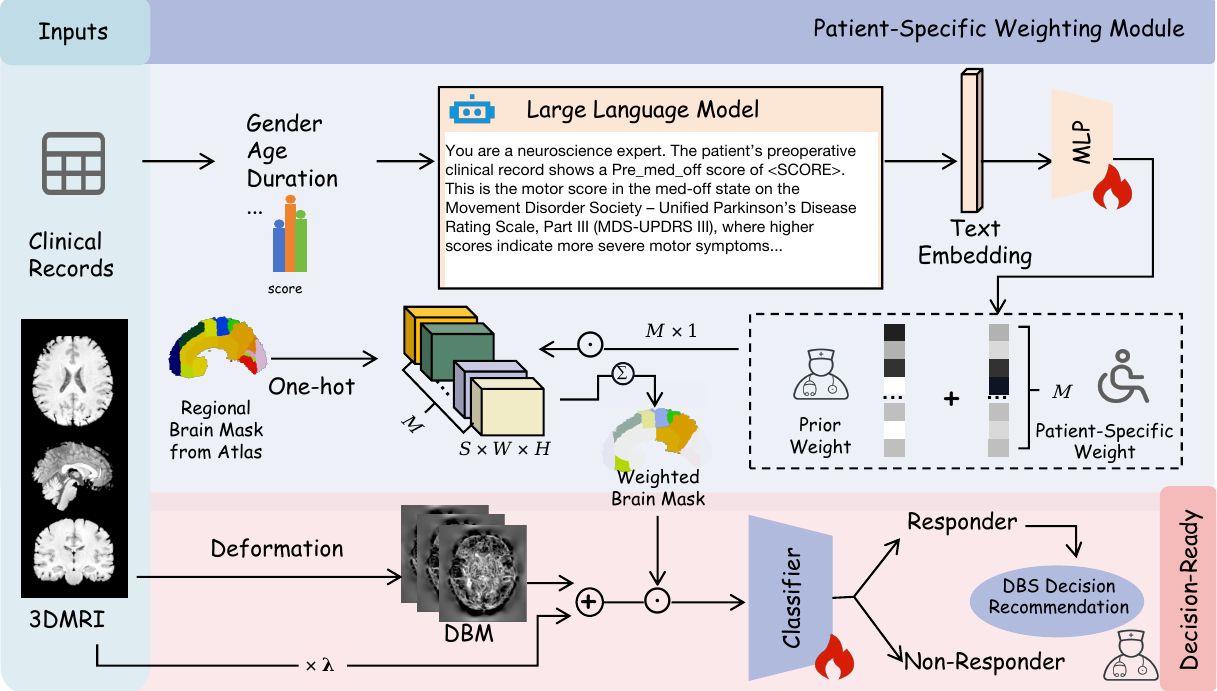}
    \caption{Overview of \textbf{PreSight}. 
Given preoperative MRI and clinical records, PreSight predicts postoperative DBS responsiveness in PD. 
Clinical variables are converted into text prompts and encoded by a large language model into patient-specific embeddings, which modulate region-prior morphometric maps via a \emph{Patient-Specific Weighting Module}. The resulting weighted deformation-based morphometry (DBM) features are fed into a classifier to determine responders and non-responders.
    }
    \label{fig:presight}
\end{figure*}

\textbf{Background and Task Definition.}
Before introducing our approach, we briefly survey the technical background relevant to predicting DBS outcomes for Parkinson’s disease and the format of the data employed in our study. We formulate pre-operative PD outcome prediction as an ordinal classification task with clinically meaningful thresholds (e.g., $<$30\%, $\geq$30\% improvement). We use pre-operative, patient-specific triplet inputs: structured clinical measures (e.g., age, sex, disease duration, baseline MDS-UPDRS-III\cite{Goetz2008MDSUPDRS}, and cognitive/affective scales), the corresponding T1w MRI, and a DBM volume computed from that MRI.

We work in a 3D template domain $\Omega \subset \mathbb{R}^3$. A population-average template $T$ was built from all pre-operative MRIs using ANTs (rigid $\to$ affine $\to$ SyN) \cite{Tustison2021ANTsX}. 
For each subject $n$, we apply bias correction, brain extraction, and resampling to the template grid, obtaining a T1 volume $I_n:\Omega\to\mathbb{R}$. A location (voxel center) in the template is denoted by
$x=(x_1,x_2,x_3)^\top \in \Omega$.
We then estimate a diffeomorphic warp $ \psi_n:\Omega\to\Omega \quad      \text{(subject}\to\text{template)} 
\quad\text{s.t.}\quad T(\psi_n(x)) \approx I_n(x)$,
and denote its inverse by $\phi_n=\psi_n^{-1}$ (template$\to$subject). 
The subject can be represented on the template grid as:
\begin{equation}
    \tilde I_n(x)=I_n(\phi_n(x)).
\end{equation}

DBM characterizes local brain–morphology differences by aligning each subject's pre-operative T1-weighted MRI to a consensus population template via high-dimensional diffeomorphic registration\cite{elias2022structuro}. Concretely, DBM computes the (log-)Jacobian of the diffeomorphic warp, where positive values indicate local expansion and negative values indicate atrophy. Because DBM analyzes geometry rather than raw intensities, it provides a contrast-agnostic, whole-brain characterization of PD-related atrophy, including within DBS-targeted movement-related brain networks, which makes it suitable as a pre-operative input for outcome modeling \cite{Zeighami2015DBM,Borghammer2010DBM,Wang2022MorphometryDBS}. Accordingly, we use DBM-derived maps as morphometric inputs to our downstream prediction model.

From the preceding registration stage, we can obtain a diffeomorphic warp between subject and template: either the subject$\rightarrow$template map $\psi_n:\Omega\to\Omega$ or its inverse $\phi_n=\psi_n^{-1}$ (template$\rightarrow$subject). Then, we compute the voxelwise (log-)Jacobian determinant from the deformation field to encode local shape change. At each 3D point $x$, the spatial Jacobian matrix of $\phi_n$ is:
\begin{equation}
\nabla\phi_n(x) \;=\;
    \begin{bmatrix}
\partial \phi_{n,1}/\partial x_1 & \partial \phi_{n,1}/\partial x_2 & \partial \phi_{n,1}/\partial x_3\\
\partial \phi_{n,2}/\partial x_1 & \partial \phi_{n,2}/\partial x_2 & \partial \phi_{n,2}/\partial x_3\\
\partial \phi_{n,3}/\partial x_1 & \partial \phi_{n,3}/\partial x_2 & \partial \phi_{n,3}/\partial x_3
\end{bmatrix}\in\mathbb{R}^{3\times 3}.
\end{equation}
The local volumetric change is quantified by its determinant
\begin{equation}
   J_n(x)=\det\big(\nabla\phi_n(x)\big),\qquad \ell J_n(x)=\log J_n(x), 
\end{equation}
where $J_n(x)>1$ indicates local expansion and $J_n(x)<1$ indicates local contraction
relative to the template.
At last, a light Gaussian smoothing is applied to reduce residual misalignment:
\begin{equation}
    \widehat{\ell J}_n=\mathcal{S}(\ell J_n).
\end{equation}

Deep Brain Stimulation (DBS) of basal ganglia targets is a standard surgical therapy for medication-refractory Parkinson’s disease. In practice, candidacy and perceived benefit are summarized by the expected change in motor severity (MDS-UPDRS~III, off-medication) at a fixed follow-up after surgery and initial programming.
Clinically, an improvement of $\geq\!30\%$ is commonly regarded as meaningfully beneficial for daily motor function and often guides surgical decision-making; however, realized outcomes are also influenced by post-operative technical factors (lead placement, programming, medication adjustments).
To avoid conflating pre-operative biology with these post-operative influences, we formulate a pre-operative prognostication task using a conservative, clinician-grounded threshold.

\textbf{Problem setup.}
For patient $n$, we are given pre-operative inputs:
a template-space T1 MRI $\tilde I_n$,
its DBM map $\widehat{\ell J}_n$,
and optional structured covariates $\mathbf{C}_n$ (age, sex, disease duration, baseline MDS-UPDRS~III, off-medication).
Let $S_{\text{pre}}$ and $S_{\text{post}}$ denote MDS-UPDRS~III (off-medication) at baseline and at a fixed post-operative follow-up, respectively.
We define the improvement rate:
\begin{equation}
\mathrm{IR} \;=\; \frac{S_{\text{pre}} - S_{\text{post}}}{S_{\text{pre}}} \;\in\; [-\infty,\,1].
\label{eq:ir}
\end{equation}
Outcome prediction is cast as a binary classification with threshold $\tau$ (default $\tau\!=\!0.30$):
\begin{equation}
y \;=\; \mathbbm{1}\!\left\{\mathrm{IR} \ge \tau\right\},
\qquad
\hat{y}_n \;=\; f_{\theta}\!\big(\tilde I_n,\, \widehat{\ell J}_n,\, \mathbf{C}_n\big)\in[0,1],
\label{eq:label_and_model}
\end{equation}
where $f_\theta$ outputs the estimated probability of achieving a clinically meaningful improvement using \emph{only} pre-operative information.

\medskip\noindent
This design explicitly targets pre-operative decision support---predicting likely benefit before surgery to inform whether to proceed with DBS.

\textbf{Patient-Specific Weighting Module.}
An overview of the pipeline is shown in Fig.~\ref{fig:presight}.
Given pre\mbox{-}operative inputs for patient $n$---a template\mbox{-}space T1w MRI 
$\tilde I_n\in\mathbb{R}^{s\times h\times w}$, its DBM map $\widehat{\ell J}_n\in\mathbb{R}^{s\times h\times w}$, 
and baseline clinical covariates $\mathbf{C}_n$, we predict a binary post\mbox{-}operative outcome 
$\hat y_n \in [0,1]$ defined in Eq.~\eqref{eq:label_and_model} (using the conventional responder threshold $\tau=0.30$), where $(s,h,w)$ are the spatial sizes.
In PD, the severity and pattern of pathology differ across regions. Treating the whole brain uniformly tends to dilute signals from critical loci and weakens outcome-relevant discrimination. We therefore perform atlas-based regional partitioning and build a voxelwise set of one-hot masks. To better capture signals associated with postoperative improvement, we then apply region-wise weights: PD-relevant regions receive higher weights, whereas less relevant regions are down-weighted, guiding the model to focus on disease-related areas. 
In detail, using the Harvard--Oxford atlas~\cite{HarvardOxfordFSL,HarvardOxfordRRID}, we partition the brain into 
$m=48$ regions. Let $M\in\{0,1\}^{m\times s\times h\times w}$ denote the voxelwise one\mbox{-}hot region tensor, 
where the first axis indexes regions. Let $\mathcal{V}=\{1,\ldots,s\}\times\{1,\ldots,h\}\times\{1,\ldots,w\}$ denote the voxel grid; we index voxels by $(i,j,k)\in\mathcal{V}$. As a result, at every voxel $\forall (i,j,k)\in\mathcal{V}$, exactly one region is active:
\begin{equation}
    \sum_{r=1}^{m} M_{r,i,j,k} \;=\; 1,
    \qquad M_{r,i,j,k}\in\{0,1\}.
\end{equation}
Furthermore, because disease course and related factors vary across patients, the regional weights should adapt to each individual. Accordingly, we introduce the PSWM, which performs dynamic region-wise input weighting by combining atlas-based priors with patient-conditioned weights.

Let $\mathbf{p}\in\mathbb{R}_{+}^{m}\ (m{=}48)$ be fixed prior region weights.
For patient n, we obtain a clinical embedding via a prompt-conditioned LLM and map it to per-region weight by MLP denoted as $g_{\phi}$:
\begin{equation}
    \mathbf{t}_n = \mathcal{G}_{\Pi}(\mathbf{C}_n)\in\mathbb{R}^{d}, 
    \qquad 
    \boldsymbol{\delta}^{(n)} = g_{\phi}(\mathbf{t}_n)\in\mathbb{R}^{m}.
\end{equation}
We then produce one scalar gate per region via a sigmoid $\sigma$:
\begin{equation}
    w^{(n)}_{r} \;=\; \sigma\!\big(\,\alpha\, p_{r} \;+\; \beta\, \delta^{(n)}_{r}\,\big)
    \;\in\; (0,1),
    \qquad r=1,\ldots,m ,
    \label{eq:sigmoid_gate}
\end{equation}
with \(\alpha,\beta>0\) controlling the influence of the prior and the personalized term.
Let the region-wise weights $w^{(n)}_{r}$ and one\mbox{-}hot masks $M$ define a voxelwise weight map
$W_n[i,j,k]=\sum_{r=1}^{m} w^{(n)}_{r} M_{r,i,j,k}$.

\textbf{Outcome Prediction.}
We form the weighted inputs and fuse the modalities via a weighted element-wise summation:
\begin{equation}
    \tilde I_n^{(\mathrm{w})} \;=\; W_n \odot \tilde I_n,\qquad
    \widehat{\ell J}_n^{(\mathrm{w})} \;=\; W_n \odot \widehat{\ell J}_n.
\end{equation}
\begin{equation}
    X_n \;=\; \lambda\,\tilde I_n^{(\mathrm{w})} \;+\; \widehat{\ell J}_n^{(\mathrm{w})} ,
\end{equation}
where $\lambda$ is a scalar hyperparameter. Note that clinical variables $C_n$ are not directly concatenated as input features to the classifier. Instead, they serve as a modulator to inform the PSWM, enabling the model to adaptively focus on patient-specific, disease-relevant brain regions.
At last, we map the weighted input $X_n$ to a scalar logit via a single model $f(\cdot;\theta)$ that comprises the image  $E_{\mathrm{img}}$, global pooling, and a linear head:
\begin{equation}
    s_n \;=\; f_{\theta}(X_n)\in\mathbb{R},\qquad 
    \hat y_n \;=\; \sigma(s_n)\in(0,1).
\end{equation}
Given labels $y_n\in\{0,1\}$, we optimize the logits-based binary cross-entropy:
\begin{equation}
    \mathcal{L}_{\mathrm{BCE}}
    \;=\; \frac{1}{N}\sum_{n=1}^{N} \Big(\log\!\big(1+\mathrm{e}^{\,s_n}\big) - y_n s_n\Big).
\end{equation}


\section{Experiments}
\subsection{Dataset and Preprocessing}
We assemble a multicenter, multimodal PD–DBS cohort for presurgical outcome prediction, comprising 366 development subjects and 34 independent external subjects. Preoperative imaging and structured clinical variables are paired with 6–12~month postoperative assessments to define responder labels. Following prior work \cite{elias2022structuro}, random downsampling is applied for class balance. All MRIs undergo standardized preprocessing (N4 correction, brain extraction, and diffeomorphic normalization to template space), resampling to \(182\times218\times182\) resolution and z-score normalization using training-cohort statistics to prevent data leakage. DBM is computed as the log-Jacobian of the deformation field. To encode anatomical priors, the Harvard–Oxford atlas is rasterized into voxelwise one-hot masks for 48 regions, enabling region-aware modeling.

\textbf{Implementation Details.}
We train the model using AdamW (initial learning rate $1\times10^{-4}$, weight decay 0.05) with 10-epoch linear warm-up followed by cosine decay, for up to 100 epochs with early stopping (batch size 4). The loss weighting coefficient $\lambda$ is set to 0.1. Clinical variables are converted into natural-language prompts and encoded using Qwen2.5-VL-Embedding. Performance is evaluated at the subject level using Accuracy (ACC), True Positive Rate (TPR), and False Positive Rate (FPR). We adopt two evaluation settings: (i) \textbf{Internal Test}, using an 8/2 patient-wise split within the development cohort without subject overlap; and (ii) \textbf{External Test}, where the trained model is directly applied to the independent cohort with identical preprocessing and fixed decision thresholds, without fine-tuning or recalibration.
\begin{figure*}[t]
  \centering
  \includegraphics[width=0.9\textwidth]{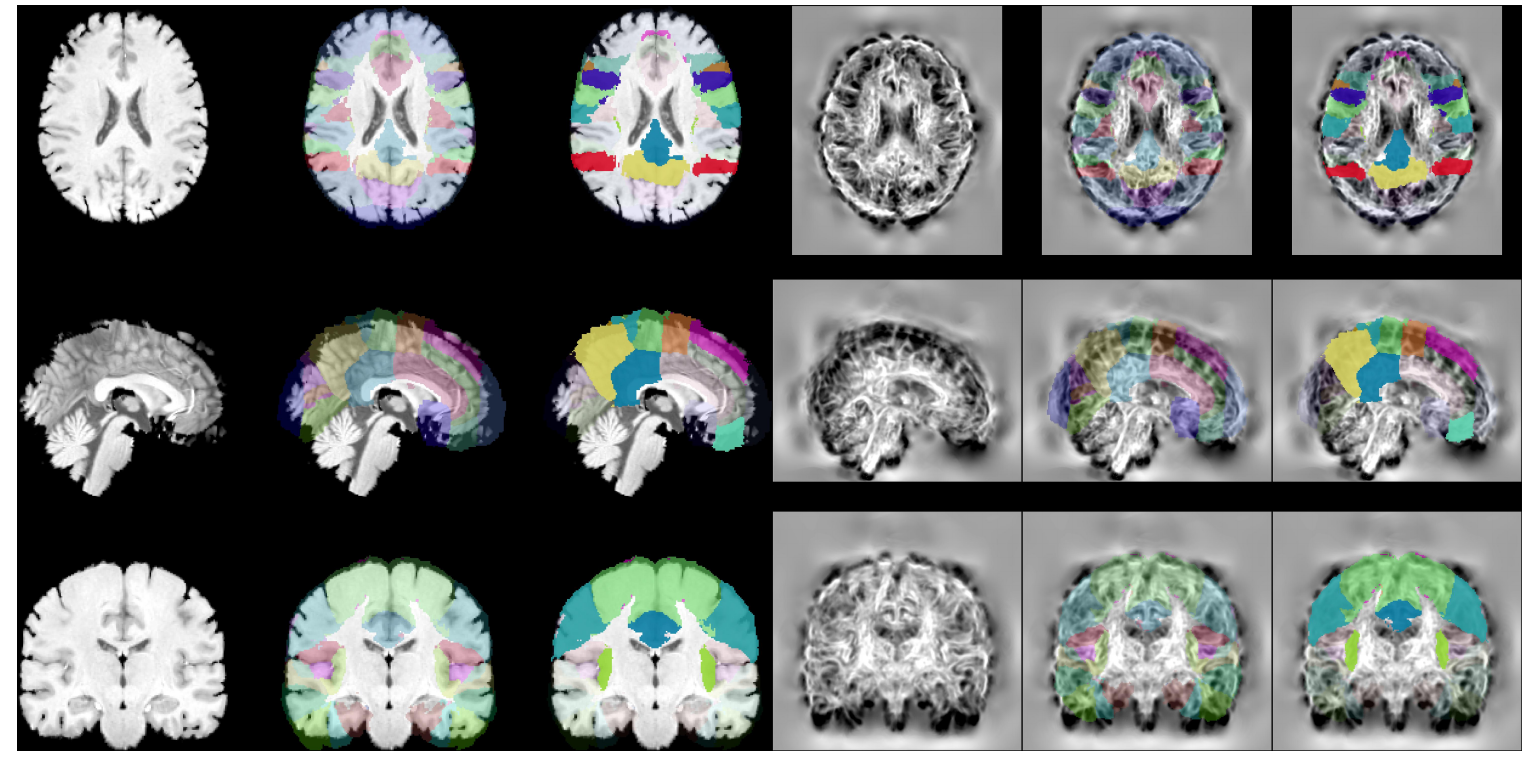}
  \caption{Multi-view visualization of parcellation and region-wise weights (axial, sagittal, coronal). 
Columns show MRI, atlas parcellation, region-wise weighting (opacity $\propto$ weight), and DBM visualizations (DBM, DBM+parcellation, DBM+weighting).
}
  \label{fig:vis}
\end{figure*}

\subsection{Comparison Results and Analysis}
We compare three baseline families under identical splits, preprocessing, optimization, and a fixed decision threshold transferred unchanged to the external cohort: (i) clinical+XGBoost, (ii) radiomics+ML (atlas-based features with LASSO and linear SVM \cite{aerts2014radiomics}), and (iii) imaging-only 3D backbones (S3D \cite{wald2025revisiting}, Swin-UNETR \cite{hatamizadeh2022swinunetr}, 3D-MAE/ViT \cite{chen2023masked}). On the internal test set, clinical-only modeling (clinical+XGBoost) achieves 63.89\% ACC with high FPR (75.00\%), while radiomics+ML improves ACC to 75.00\% (best SVM) at the cost of reduced sensitivity. Imaging-only deep models further enhance discrimination (best: S3D, 80.56\% ACC). In contrast, PreSight reaches 88.89\% ACC with 95.83\% TPR and 16.67\% FPR, yielding an absolute gain of 8.33 points over the strongest imaging-only baseline while matching the highest sensitivity and lowest FPR. On the external cohort, PreSight maintains 85.29\% ACC (94.12\%/23.53\% TPR/FPR), outperforming S3D by +2.94 ACC and +5.88 TPR at comparable FPR. Clinical-only and radiomics+ML degrade substantially out-of-distribution, whereas PreSight shows only a 3.60-point internal-to-external ACC drop. 
These results indicate that neither tabular nor T1w-only models are sufficiently discriminative. Incorporating DBM and patient-specific regional weighting enables more consistent sensitivity–specificity trade-offs and improved generalization.
We further visualize regional contributions in Fig.~\ref{fig:vis}. 
Using atlas priors with patient-specific weighting, prominent weights emerge in the Postcentral Gyrus (SMN), Precuneous Cortex (DMN), and Superior Parietal Lobule (DAN). 
This concentration across sensorimotor, default-mode, and dorsal attention networks is consistent with reported PD-related motor and higher-order network dysfunction \cite{burciu2018imaging,gao2016study}.

\begin{table*}[t]

\footnotesize
\setlength{\tabcolsep}{6pt}
\renewcommand{\arraystretch}{1.15}
\centering

\begin{threeparttable}
\caption{Comparison across input modalities and backbones. Higher is better for ACC/TPR (↑); lower is better for FPR (↓). All methods use identical patient-wise splits, preprocessing, and fixed decision thresholds chosen on the internal development split and transferred to the external cohort without fine-tuning or recalibration.}
\label{tab: Comparison with SOTAs}

\begin{tabularx}{\textwidth}{
    p{0.30\textwidth}  
    CCCCCC             
}
\toprule
\multirow{2}{*}{\textbf{Method}}   &
\multicolumn{3}{c|}{\textbf{Internal Test}} & \multicolumn{3}{c}{\textbf{External Test}}\\
\cmidrule(lr){2-4}\cmidrule(lr){5-7}
&  \textbf{ACC (\textuparrow)} & \textbf{TPR (\textuparrow)} & \textbf{FPR(\textdownarrow)} & \textbf{ACC (\textuparrow)} & \textbf{TPR (\textuparrow)} & \textbf{FPR(\textdownarrow)} \\
\midrule
Clinical+XGBoost\cite{chen2016xgboost}        & 63.89 & 83.33& 75.00 & 55.88 & 52.94 & 41.18 \\
Radiomics + LASSO \cite{aerts2014radiomics}           & 69.44 & 79.17 & 50.00 & 58.82 & 64.71 & 47.06 \\
Radiomics + SVM   \cite{aerts2014radiomics}& 75.00 & 70.83 & 16.67 & 61.76 &76.47  & 52.94\\

S3D \cite{wald2025revisiting}                 & 80.56 & 91.67 & 41.67 & 82.35 & 88.24 & 23.53 \\
Swin UNETR \cite{tang2022self}      & 77.78& 95.83 & 58.33& 73.53 & 82.35  & 35.29\\
3D-MAE/ViT \cite{chen2023masked}     & 72.22 & 91.67 & 66.67 & 76.47 & 88.24  &35.29 \\
\rowcolor{gray!20}
\textbf{PreSight} & \textbf{88.89} & \textbf{95.83} & \textbf{16.67} & \textbf{85.29} & \textbf{94.12} & \textbf{23.53} \\
\bottomrule
\end{tabularx}
\end{threeparttable}
\end{table*}
\textbf{Ablation Studies.} 
We isolate the effects of DBM and PSWM. 
DBM-only yields 77.78\% ACC. Adding fixed regional priors improves performance to 80.56\% (+2.78 pp), indicating that anatomically informed weighting mitigates whole-brain averaging. 
Using patient-specific offsets instead achieves 83.33\% (+5.55 pp over DBM-only), showing stronger gains from individualized reweighting. 
Combining priors and patient-specific weighting attains the best result (88.89\%), confirming their complementarity. 
In contrast, removing DBM while retaining PSWM reduces accuracy to 72.22\%, demonstrating that explicit morphometry is essential and that PSWM is most effective when coupled with DBM.
\begin{table}[t]
\centering
\footnotesize
\setlength{\tabcolsep}{4pt}
\renewcommand{\arraystretch}{1}
\centering
\caption{Ablation on DBM and the Patient-Specific Weighting Module (PSWM). 
\checkmark indicates the component is enabled. 
}
\begin{tabular}{c|cc|c}
\toprule
\textbf{DBM} &
\multicolumn{2}{c|}{\textbf{PSWM components}} &
\textbf{ACC} \\
 & \textbf{Prior } & \textbf{Patient-specific} & (\%) \\
\midrule
\checkmark &             &             & 77.78 \\
\checkmark & \checkmark  &             & 80.56 \\
\checkmark &             & \checkmark  & 83.33 \\
\checkmark & \checkmark  & \checkmark  & \textbf{88.89} \\
           & \checkmark  & \checkmark  & 72.22 \\
\bottomrule
\end{tabular}
\end{table}

\section{Conclusion}
We present PreSight, a presurgical model that integrates DBM and T1w MRI through a patient-specific regional weighting module conditioned on clinical embeddings. Under a standardized protocol with fixed external thresholds, PreSight achieves 88.89\% internal and 85.08\% external accuracy with favorable TPR/FPR trade-offs and calibrated outputs. 
Ablations confirm that DBM provides the essential structural foundation, while patient-adaptive weighting yields complementary gains, with qualitative alignment to disease-relevant circuits. Together, these results demonstrate robust and transferable prediction of DBS responsiveness.
\bibliographystyle{unsrt}
\bibliography{main}

%
%
%






\end{document}